\documentclass[a4paper,conference]{template/IEEEtran}

\ifCLASSINFOpdf
\else
\fi
\usepackage{amsmath}
\usepackage{graphicx}
\usepackage{subcaption}
\usepackage[dvipsnames]{xcolor}
\usepackage{subcaption}
\usepackage{multirow, makecell}
\usepackage{tabularx} % table with even-width columns
\usepackage{wrapfig} % Wrap text around figure
\usepackage{pgfplots} % plots with altex fonts
\usepackage{layouts} % Used for \printinunitsof{in}\prntlen{\textwidth}

\usepackage{ifthen}
\newcommand\hide[1]{}
\newboolean{show_todos}
\newboolean{show_notes}
\newboolean{dont_show_bool}
\setboolean{show_todos}{true}%{false}%
\setboolean{show_notes}{false}%
\setboolean{dont_show_bool}{false}%{true}%
\newcommand\todo[1]{\ifthenelse{\boolean{show_todos}}{\textcolor{red}{\textbf{ToDo: }#1}}{\hide{#1}}}
\newcommand\note[1]{\ifthenelse{\boolean{show_notes}}{\textcolor{blue}{#1}}{\hide{#1}}}
\newcommand\dontshow[1]{\ifthenelse{\boolean{dont_show_bool}}{\hide{#1}}{\textcolor{blue}{#1}}}
\newcommand\laura[1]{\ifthenelse{\boolean{show_notes}}{\textcolor{orange}{\textbf{Laura:} #1}}{\hide{#1}}}
\newcommand\tim[1]{\ifthenelse{\boolean{show_notes}}{\textcolor{orange}{\textbf{Tim:} #1}}{\hide{#1}}}
\begin{document}

\title{Street-Map Based Validation of Semantic Segmentation in Autonomous Driving}

% \author{
% \IEEEauthorblockN{Laura von Rueden, Tim Wirtz}
% \IEEEauthorblockA{Fraunhofer Center for Machine Learning\\
% Fraunhofer IAIS, Germany\\
% laura.von.rueden@iais.fraunhofer.de}
% \and
% \IEEEauthorblockN{Fabian Hueger, Jan David Schneider}
% \IEEEauthorblockA{Volkswagen Group Automation\\
% Wolfsburg, Germany}
% \and
% \IEEEauthorblockN{Christian Bauckhage}
% \IEEEauthorblockA{Fraunhofer Center for Machine Learning\\
% Fraunhofer IAIS, Germany}
% }



\author{
\IEEEauthorblockN{
Laura von Rueden\IEEEauthorrefmark{1}\IEEEauthorrefmark{3},
Tim Wirtz\IEEEauthorrefmark{1},
Fabian Hueger\IEEEauthorrefmark{2}, 
Jan David Schneider\IEEEauthorrefmark{2},
Nico Piatkowski\IEEEauthorrefmark{1},
Christian Bauckhage\IEEEauthorrefmark{1}
}
\IEEEauthorblockA{
\IEEEauthorrefmark{1}
Fraunhofer Center for Machine Learning, Fraunhofer IAIS, Sankt Augustin, Germany
}
\IEEEauthorblockA{
\IEEEauthorrefmark{2}
Volkswagen Group Automation, Wolfsburg, Germany
}
\IEEEauthorblockA{
\IEEEauthorrefmark{3}
laura.von.rueden@iais.fraunhofer.de
}
}

\IEEEspecialpapernotice{(Preprint - Final version will be published at IEEE)}

\maketitle

\begin{abstract}
Artificial intelligence for autonomous driving must meet strict requirements on safety and robustness, which motivates the thorough validation of learned models.
However, current validation approaches mostly require ground truth data and are thus both cost-intensive and limited in their applicability.
We propose to overcome these limitations by a model agnostic validation using a-priori knowledge from street maps.
In particular, we show how to validate semantic segmentation masks and demonstrate the potential of our approach using OpenStreetMap.
We introduce validation metrics that indicate false positive or negative road segments.
Besides the validation approach, we present a method to correct the vehicle's GPS position so that a more accurate localization can be used for the street-map based validation.
Lastly, we present quantitative results on the Cityscapes dataset indicating that our validation approach can indeed uncover errors in semantic segmentation masks.
\end{abstract}


% \fi
\IEEEpeerreviewmaketitle

% The textwidth: \printinunitsof{in}\prntlen{\textwidth}
%% Results: 7.1413in
% The columnwidth: \printinunitsof{in}\prntlen{\columnwidth}
%% Results: 3.48761in

\section{Introduction}
\label{sec:intro}

% --- Intro: Semantic segmentaion and deep learning for autonomous vehicles
Environmental perception is important for autonomous vehicles in order to assess the surrounding traffic scene and understand its context~\cite{campbell2010autonomous, pendleton2017perception}.
A key component is semantic segmentation,
which assigns pixel-wise pre-defined class labels to the input images from vehicle's cameras.
Current algorithms use machine and deep learning techniques to build models that predict semantic segments and surpass classic computer vision techniques in terms of performance~\cite{garcia2018survey, feng2019deep}.

% --- Challenge: Robust & Safe AI
The development of artificial intelligence systems brings certain challenges, especially when they are applied in safety-critical areas.
Building deep neural networks that generalize well and are robust often comes with the need for large amounts of ground truth data, which is typically acquired in expensive manual labelling processes.
To ensure the safety of AI-based systems, mechanisms that support a trustworthy development like interpretability, auditing and risk assessment are discussed with growing interest~\cite{brundage2020toward}.

% --- Importance of: Validating ML models in autonomous driving
The validation of machine learning models is particularly important in the area of highly automated driving, for example the identification and mitigation of risks of potential functional insufficiencies in neural networks used for perception~\cite{burton2017making}.
Since the perception component is responsible for the first assessment of the vehicle's surroundings, the detection and reduction of errors in this component can increase the reliability of the resulting environment model.
Proposed approaches for mitigation are the detection of prediction uncertainties~\cite{kendall2017uncertainties} and the estimation of an according error propagation~\cite{mcallister2017concrete}.

\begin{figure}[t]
\centering
\includegraphics[width=\columnwidth]{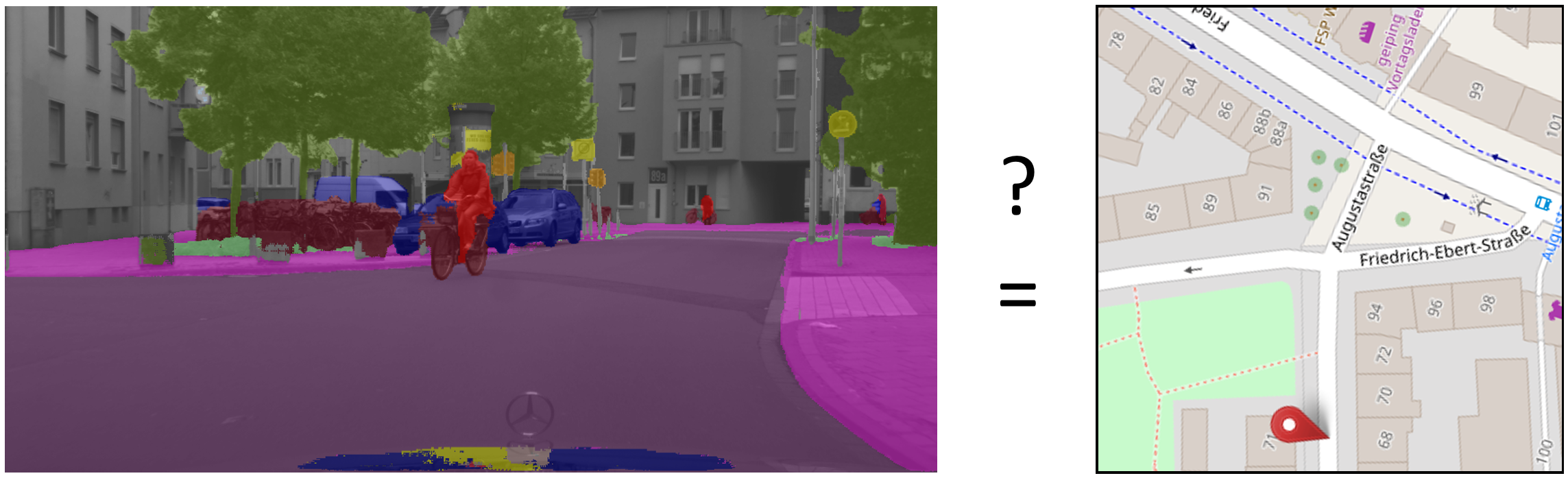}
\caption{
\textbf{Research question.} Can predicted semantic segmentation masks be validated with a-priori knowledge from street maps? The left image shows a segmentation of a traffic scene in the Cityscapes dataset~\cite{cordts2016cityscapes} and the right image shows the corresponding map~\cite{OpenStreetMap}.
Here, an intersection to the right, which is shown in the map, is not reflected in the segmentation.
}
\label{fig:intro}
\end{figure}

% --- Problem: Semantic Segmentation Errors in Drivable ares
Although state-of-the-art neural networks for semantic segmentation achieve promising results, it can still be observed that certain objects of the drivable area are not detected correctly. Moreover, smaller networks being used for embedded purposes are often comparatively less accurate than state-of-the-art networks with arbitrary size. As an example, roads and pedestrian walks could be mixed up in difficult lighting conditions or unusual terrain
like in the segmentation in Figure~\ref{fig:intro}.

% --- Approach: Knowledge injection, Segmentation Validation by Street Maps
We propose to support the goal of safe artificial intelligence in autonomous driving by applying the idea of informed machine learning~\cite{vonrueden2020informed} and validate learned models with a-priori knowledge.
In this paper, we suggest to compare semantic segmentation masks to the structured semantic information in street maps, as illustrated in Figure~\ref{fig:intro}, and present a novel method that computes the overlap of drivable area between the segmentation output and the map.
% Approach inspiration
Our approach is inspired by how human drivers would perceive environments: When they find themselves in a new environment, they often consult external knowledge sources such as street maps and compare what they see in their vicinity to what they see on the map.

\begin{figure*}[t]
\centering
\includegraphics[width=\textwidth]{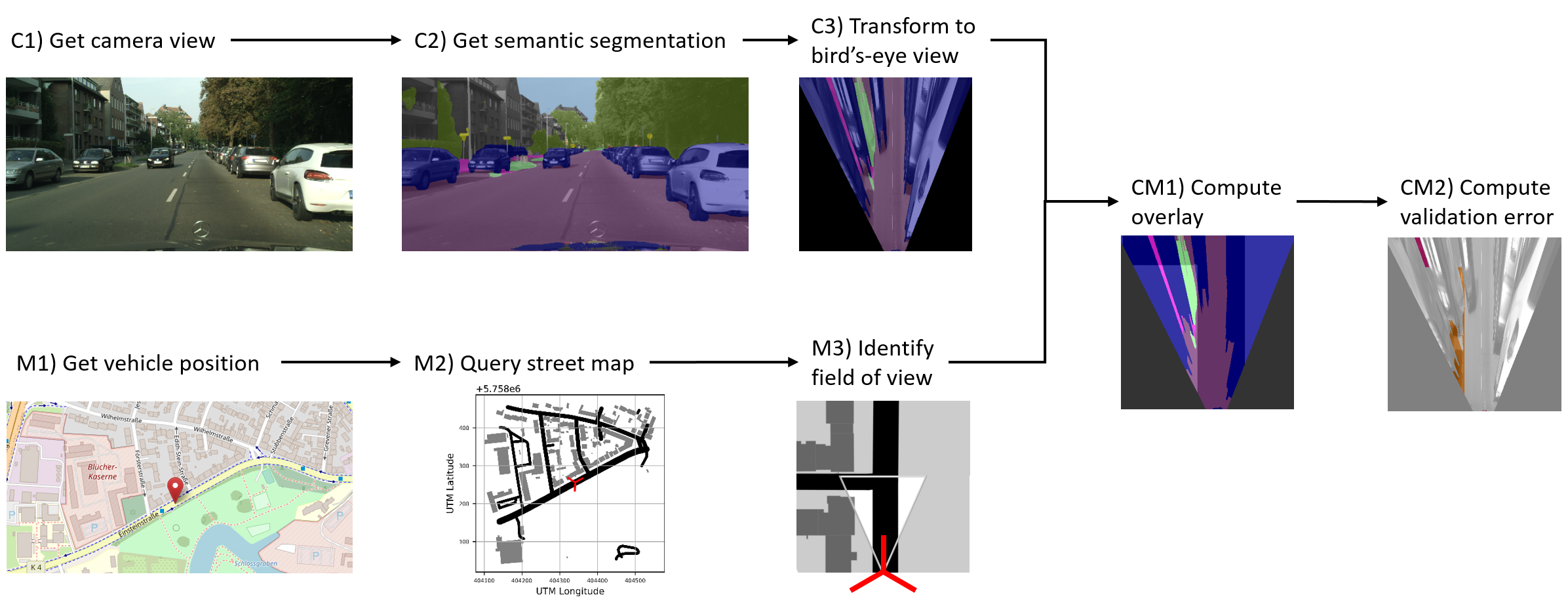}
\caption{
\textbf{Approach overview}: We validate the drivable area in semantic segmentation masks with a-priori knowledge from street maps. For this, we combine the segmentation mask of a given camera view with the street map that corresponds the given vehicle position.
We compute an overlay of the segmentation in bird's-eye view and the street map's field of view, omitting occluding, dynamic objects such as other vehicles or vegetation.
The overlay is used to identify validation error regions, which we classify into detected false positives (indicated by a low $IoS$; visualized with orange red pixels) and false negatives (indicated by a low $IoM$; visualized with pink red pixels).
The steps with \textit{C} describe tasks involving a \textit{c}amera view and the steps with \textit{M} describe tasks involving a \textit{m}ap view.
}
\label{fig:approach}
\end{figure*}

% --- Related Work
Related work comprises approaches for multi-modal perception for autonomous driving, combining the inputs from various driving data~\cite{feng2019deep}.
The combination of camera inputs and street maps for semantic segmentation has already been used to assign geographical addresses to detect buildings~\cite{ardeshir2015geo}, or to build conditional random fields for scene understanding~\cite{wang2015holistic, diaz2016lifting}.
The integration of general geographic or geometric a-priori knowledge in perception tasks has been investigated in versatile forms, for example as shape priors for object localization~\cite{murthy2017shape}, temporal priors for revisited locations~\cite{schroeder2019using}, or spatial relation graphs for object detection~\cite{xu2019spatial}.
However, to the best of our knowledge, there is not yet an approach that uses street maps for a validation of semantic segmentation masks.

% --- Approach advantages
An advantage of our proposed method is that it facilitates a model-agnostic validation of
segmentation masks. It can be applied to predictions from 
deep learning approaches, but also to traffic scene segmentations from any other approach including the ground truth data generation process itself.

Furthermore, our validation method can be applied either offline in the testing phase of learned models, or even online to assess potentials errors within the current prediction, since the approach does not require ground truth to be available.
Another advantage is that it can be used to test models even in geographical regions that had not been represented in the training data.
This is relevant because the characteristics of semantic concepts such as roads, cars, or vegetation can be very diverse across different regions, but training datasets do not always reflect this domain variety~\cite{tsai2018learning, wang2020train}.
For autonomous driving, external data sources such as street maps thus provide a valuable alternative information source for static objects present in ground truth data.

Our paper presents four contributions.
First, we introduce the approach to validate the drivable area in semantic segmentation masks using a-priori knowledge obtained from street maps, which can be used to identify prediction errors.
Second, we define new validation metrics that can be used for comparing semantic segmentations of traffic scenes to street maps.
Third, we present an algorithm for localization correction that can be used to calibrate the street map position according to the ground truth segmentation.
Fourth, we present experimental results on the Cityscapes dataset, which demonstrate that our approach can identify similar prediction errors as a validation by ground truth data.
The paper is structured accordingly.
\section{Approach: Segmentation validation}
\label{sec:approach}

Our approach validates the drivable area in a given semantic segmentation mask using the corresponding geometric structures in a given street map.
In this section we present how we combine the segmentation and the street map in an overlay. We define the validation metrics that we use for the identification of potential validation errors. Finally, we demonstrate our approach with two examples.

\subsection{Overlay of segmentation and street map}

An overview of our approach is illustrated in Figure~\ref{fig:approach}. In the following we shortly describe each step within the approach.

\paragraph*{Step C1) Get camera view}
We retrieve an image from the vehicle's front view of a traffic scene. Here we use the Cityscapes~\cite{cordts2016cityscapes} dataset.

\paragraph*{Step C2) Get semantic segmentation}
Using a neural network, we obtain a segmentation mask that maps each image pixel to a set of pre-defined class labels. In the example image in Figure~\ref{fig:approach}, the labels are visualized in a chosen color coding: \textit{road} is violet, \textit{car} is blue, \textit{pedestrian walk} is pink, etc. Here, we used a model trained by the ERFNet encoder-decoder architecture~\cite{romera2017erfnet} to create the predictions.

\paragraph*{Step C3) Transform to bird's-eye view}
To prepare the validation of the drivable area segments using the street map, we transform the segmentation image into a bird's-eye view, which corresponds to the view space of the street map.

\paragraph*{Step M1) Get vehicle position}
We get the position by reading the GPS coordinates of the vehicle and thus retrieve latitude, longitude and the heading. These values are given in the Cityscapes dataset for each camera image.
Since the accuracy of the GPS position can be a challenge, we develop a localization correction algorithm, which is further described in Section~\ref{sec:gpscorr}, and apply it to alleviate inaccuracies in the position.

\paragraph*{Step M2) Query street map graph}
For the given latitude and longitude we get the street map graph for the surrounding area. For our analysis we use data from OpenStreetMap~\cite{OpenStreetMap}, because this source offers a freely available option with sufficient data coverage
for an experimental demonstration of our approach.
For future application in production other map providers that offer more detailed and accurate information, for example in high-definition maps, might be preferable.

\paragraph*{Step M3) Identify field of view}
We transform the street map graph to an image and rotate it in the direction of the vehicle's heading, which is obtained from the metadata of the camera image. We zoom in so that it corresponds to the potential field of view from the camera mounted on the car.

\paragraph*{Step CM1) Compute overlay}
We combine the prepared images in an overlay of the semantic segmentation in bird's eye view with the road from the map image. As shown in Figure~\ref{fig:approach}, the road is illustrated as a transparent black area. This allows us to recognize the overlap between the predicted road segments from the semantic segmentation mask and the street map.

\paragraph*{Step CM2) Compute validation error}
Finally we compute the regions where the predictions of the trained model deviate from the a-priori knowledge contained in the map.
Two types of validation errors can be derived: False positive regions, i.e., where the segmentation shows a road, but the map does \textit{not} (here visualized by orange red), and false negative regions, i.e., where the segmentation does \textit{not} show a road, but the map does (visualized by pink red). For computing reliable error regions, we omit pixels that are assigned to labels that could be occluding the drivable area like, e.g., vegetation or cars.

\subsection{Validation metrics}
\label{sec:validationmetrics}

\begin{figure}[t]
    \begin{subfigure}[l]{0.49\columnwidth}
    \centering
    \includegraphics[width=0.8\columnwidth]{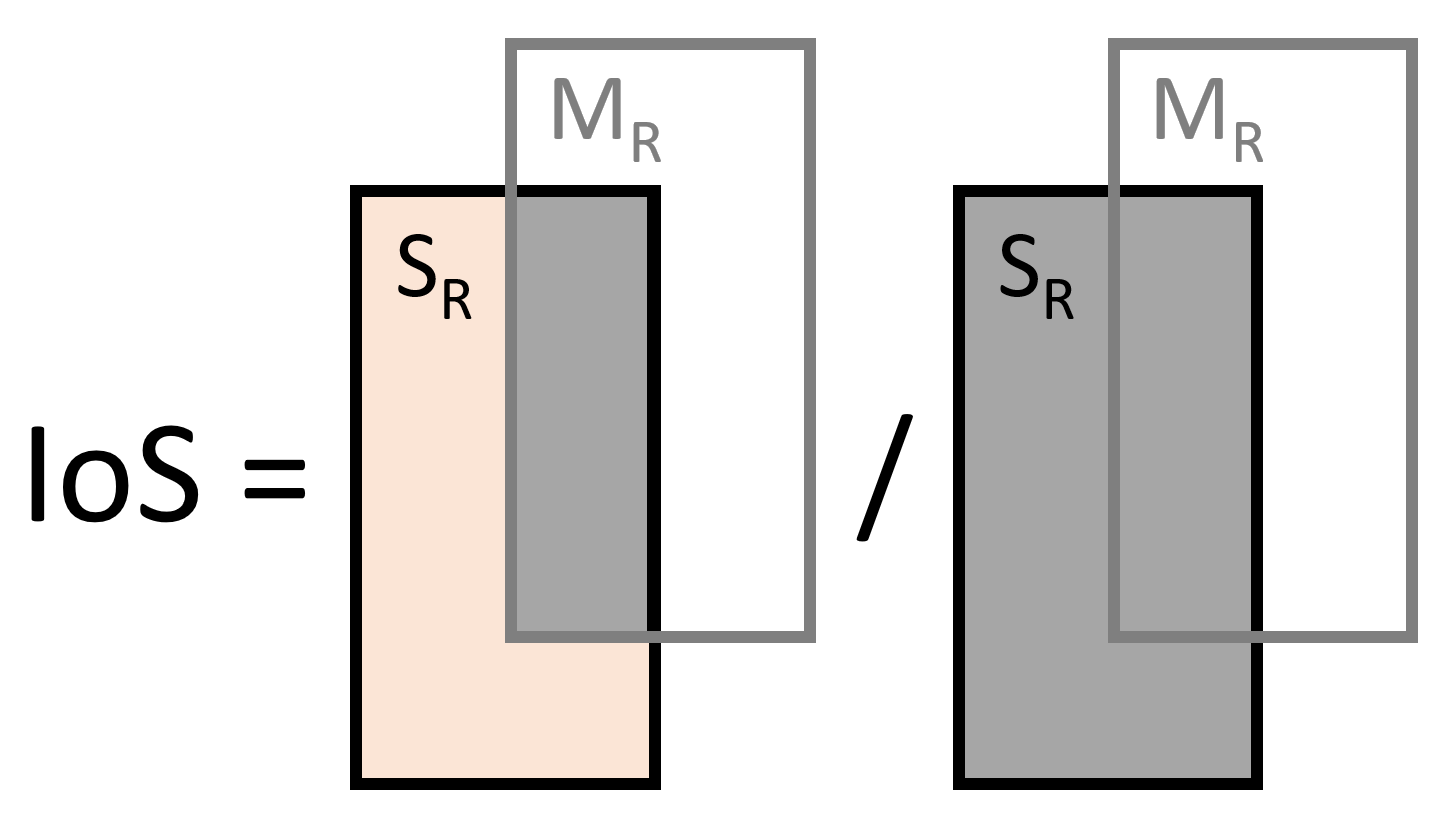}
    \subcaption{
    Intersection over Segment.
    }
    \label{fig:ios}
    \end{subfigure}
    \begin{subfigure}[r]{0.49\columnwidth}
    \centering
    \includegraphics[width=0.8\columnwidth]{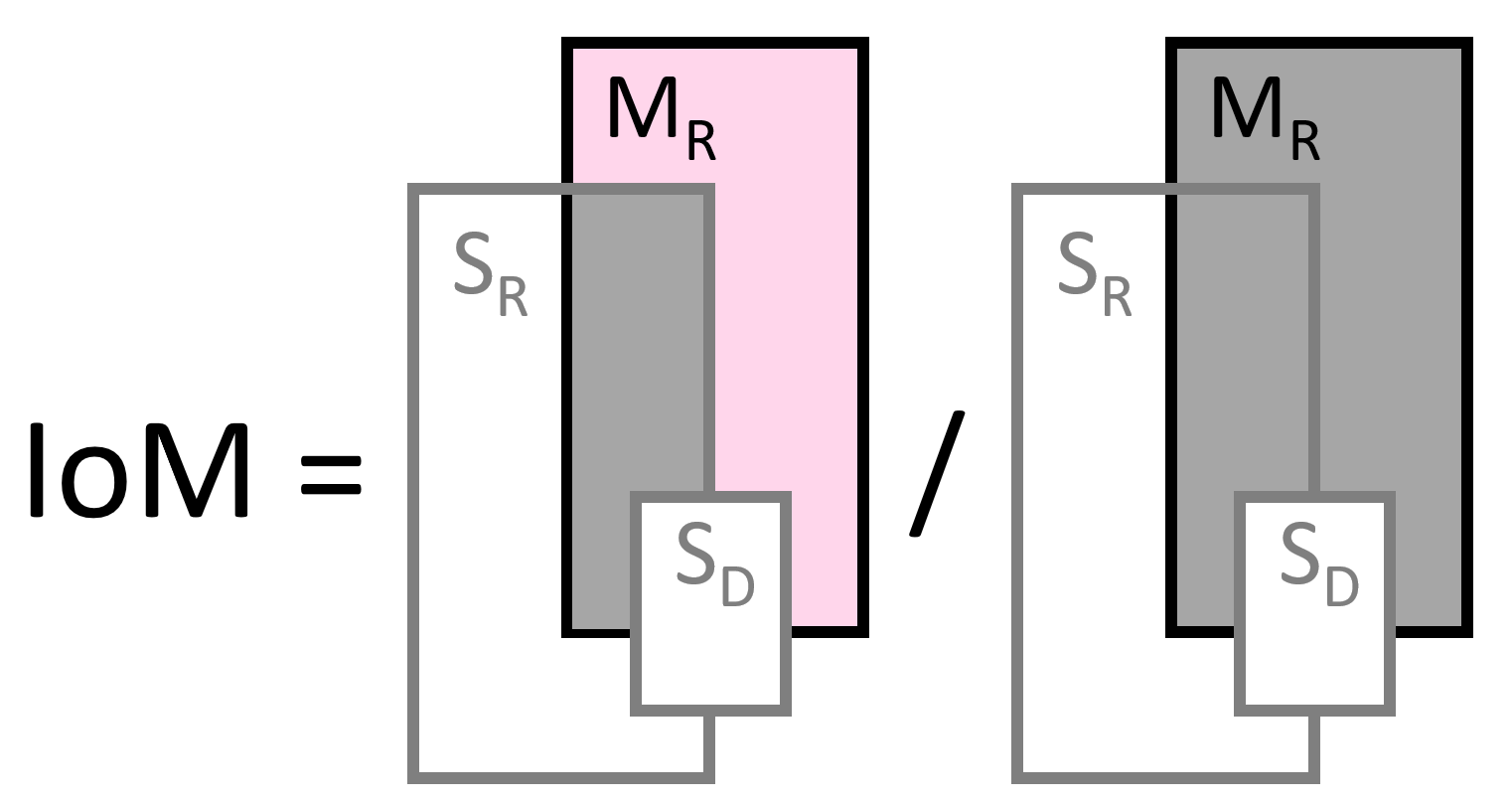}
    \subcaption{
    Intersection over Map
    }
    \label{fig:iom}
    \end{subfigure}
\caption{
\textbf{Validation metrics.}
The consistency of the semantic segmentation with the street map can be quantified by the two validation metrics that we introduced. 
a) The Intersection over Segmentation ($IoS$) quantifies the overlap of the segmentation with the map. A low $IoS$ is an indicator for false positive road segments (see orange red region).
b) The Intersection over Map ($IoM$) quantifies, vice versa, the overlap of the map with the segmentation. A low $IoM$ is an indicator for false negative road segments (see pink red region).}
\end{figure}

To quantify the overlap of semantic road segments with the street map, we introduce two new validation metrics that we call \textit{Intersection over Segmentation} ($IoS$) and \textit{Intersection over Map} ($IoM$).
In semantic segmentation, the most commonly used evaluation metric for quantifying the model performance is the \textit{Intersection over Union (IoU)}, which is defined by the area of overlap between predicted and ground truth segments divided by the area of union of both segments.
However, in our approach we are especially interested in identifying either false positive or false negative road prediction errors.
Our introduced metrics evaluate these errors for the road segments, so that they should precisely be called $IoS_R$ and $IoM_R$. For simplicity we omit this subscript and just call them $IoS$ and $IoM$ in this paper.

The $IoS$ metric quantifies the share of the semantic road segments that are covered by the roads in the map and thus helps to identify false positive errors.
Here, we define false positives as semantic road segments, where the street map does not show a road segment.
As illustrated in Figure~\ref{fig:ios}, we compute the $IoS$ as the overlap area from the semantic road segment $S_R$ and the road in the map $M_R$, divided by the area of the semantic road segment $S_R$ itself.
If the $IoS$ is high, the segmentation and street map are mostly consistent, but the lower the $IoS$, the more false positive ($FP$) pixels are in the segmentation.
Thus, the $IoS$ can also be described as the share of true positive road segments ($TP$) of all road segments:
\begin{align*}
    IoS &= \frac{S_R \cap M_R}{S_R} = \frac{TP}{TP + FP}
\end{align*}

Vice versa, the $IoM$ metric quantifies the share of the road segments in the map that are covered by the road segments in the segments and thus helps to identify false negative errors.
We define false negative errors as road segments that are not represented in the segmentation although they are present in the map.
For its calculation, any dynamic semantic segments, such as vehicles or pedestrians, and vegetation that might occlude the road segment, are omitted.
As illustrated in Figure~\ref{fig:iom}, we compute the $IoM$ as the intersection of the semantic road segment $S_R$ and the road in the map $M_R$, divided by the area of the road in the map $M_R$ minus intersections with dynamic semantic segments $S_D$.
The lower the $IoM$, the more false negative ($FN$) pixels are in the segmentation.
The $IoM$ can also be described as the share of true positive road segments from all road segments in the map:
\begin{align*}
    IoM &= \frac{S_R \cap M_R}{M_R - (M_R \cap S_D)} = \frac{TP}{TP + FN}
\end{align*}

Another metric that combines the $IoS$ and $IoM$ is the dice coefficient. It can be used to quantify the general overlap between the road in the map and in the semantic segmentation. We use it in our localization correction method, as further described in Section~\ref{sec:gpscorr}, and in our extended experiments as an initial estimation if a segmentation mask contains potential errors, as further explained in Section~\ref{sec:experiments}.

\begin{align*}
    dice &= \frac{2 \cdot(S_R \cap M_R)}{S_R + M_R - (M_R \cap S_D)} = \frac{2 \cdot TP}{2 \cdot TP + FP + FN}
\end{align*}

\begin{figure}[t]
    \centering
    \includegraphics[width=\columnwidth]{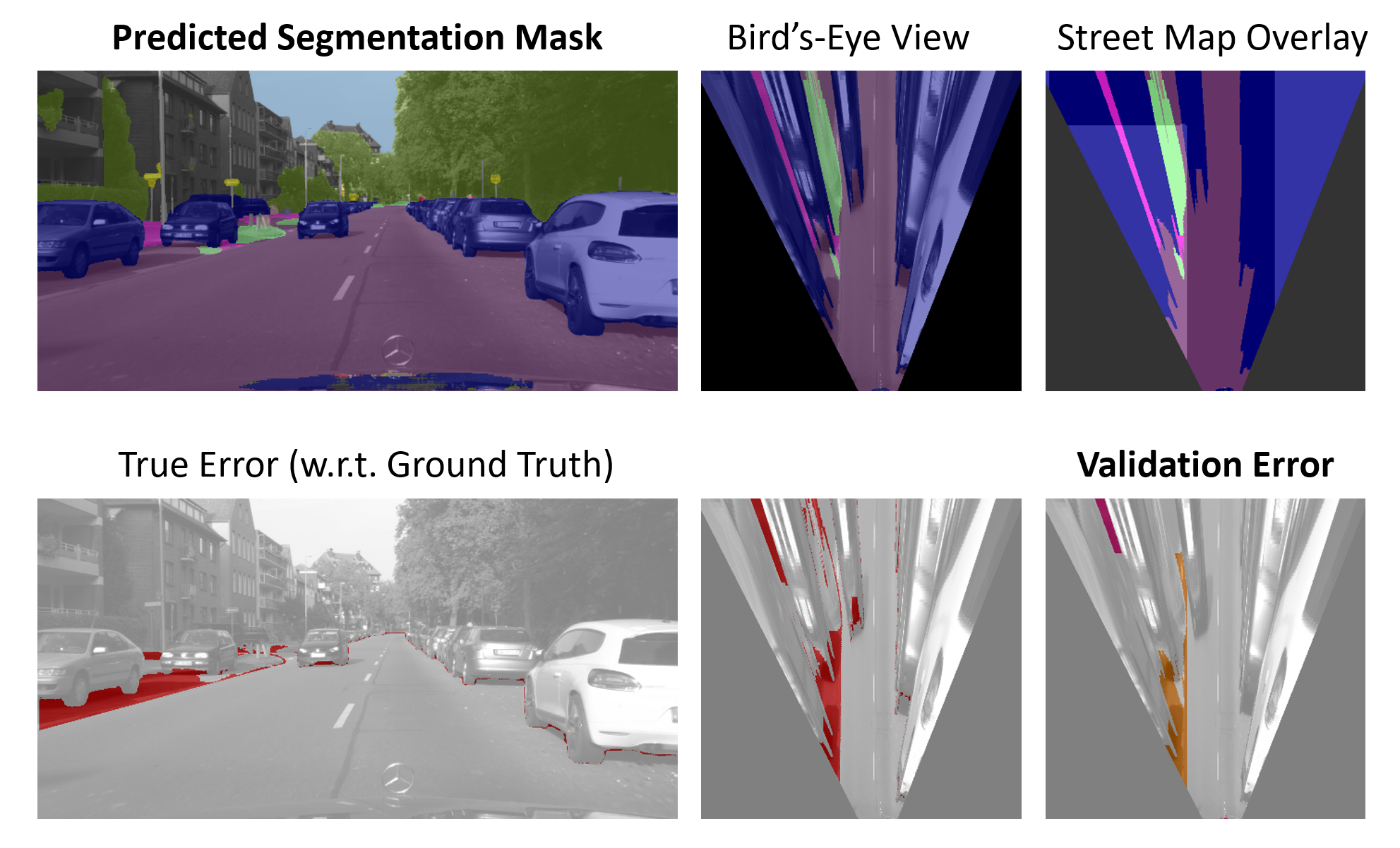}
    \caption{
    \textbf{Example for the detection of a false positive road.}
    The predicted segmentation shows a road straight forward and below the cars parked at the left side of the street. According to the ground truth there is a parking space below that parking cars.
    Our map-based validation approach identifies this deviation: The street map suggests a less broad road than in the prediction, resulting in the detection of a false positive region (see orange red color at the left side of the validation error image).
    For this image the validation metrics are $IoS = 88.03\%$, and $IoM = 97.22\%$, also reflecting the false positive road prediction.
    }
    \label{fig:ex_fp}
\end{figure}

\begin{figure}[t]
    \centering
    \includegraphics[width=\columnwidth]{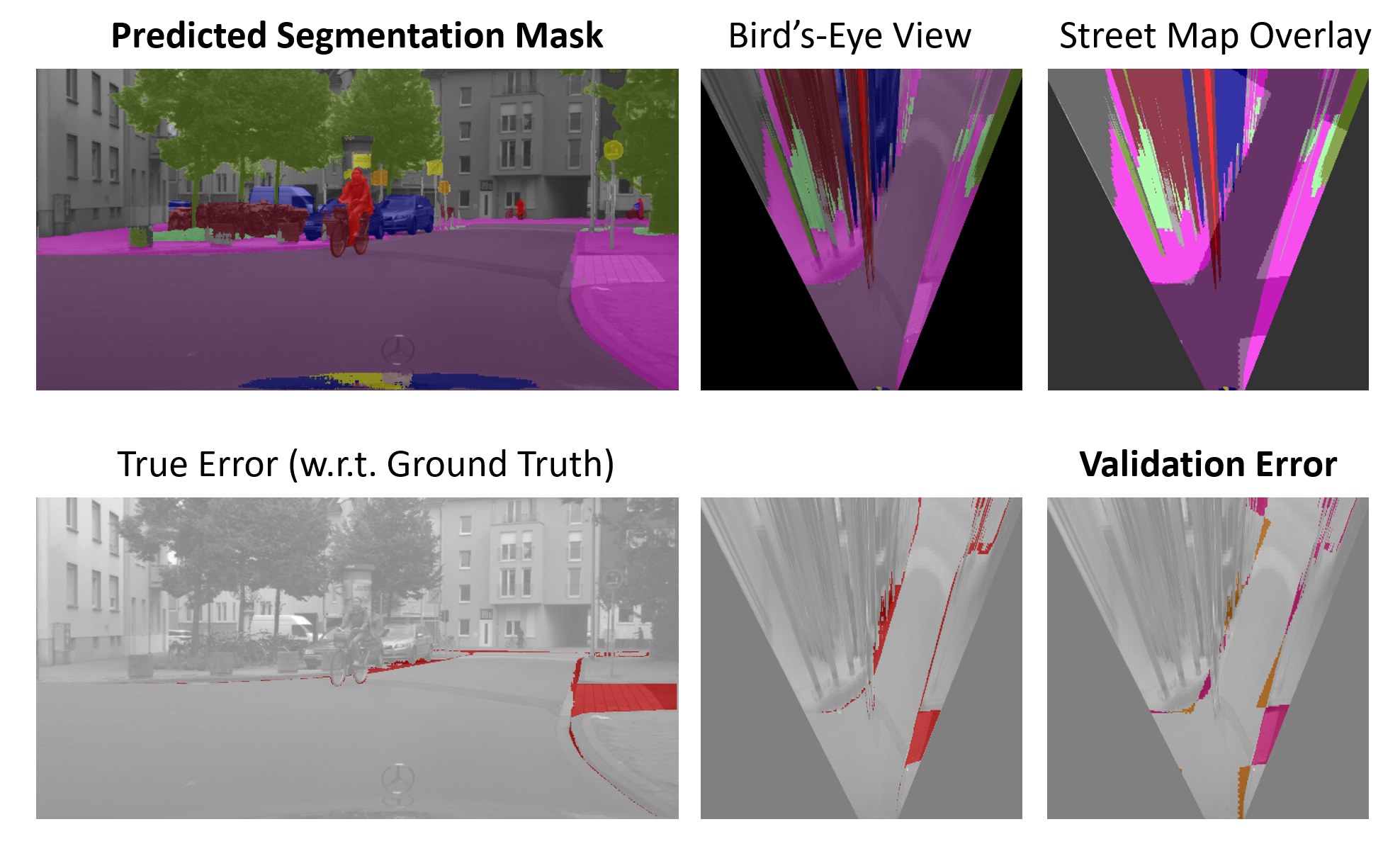}
    \caption{
    \textbf{Example for the detection of a false negative road.}
    The predicted segmentation shows a road running straight forward, although there is an intersection to the right according to the ground truth.
    Our approach identifies this deviation, too: The street map shows an intersection to the right. This results in a detected false negative region (see pink red color at the right side of the error image).
    For this image the validation metrics are $IoS = 94.27\%$, and $IoM = 90.33\%$, also reflecting the false negative road prediction.
    }
    \label{fig:ex_fn}
\end{figure}

\subsection{Example results}

We demonstrate our approach with examples for two different traffic scenes.
The first example, given in Figures~\ref{fig:ex_fp}, shows a traffic scene where the ground truth semantic segmentation shows a parking space at the left side, but the prediction shows a road.
Our approach identifies this false positive road segment, also indicated by a lower $IoS$ metric. 
The second example, given in Figure~\ref{fig:ex_fn}, shows a scene where the ground truth semantic segmentation shows a road intersection to the right, but the prediction does not.
Again, our approach identifies this false negative, also supported by a lower $IoM$ metric.
%\newpage\newpage
\section{Approach: Localization correction}
\label{sec:gpscorr}

% --- Motivation: Challenge of localization precision
The validation of segmentation masks using street map data poses a challenge with respect to the localization precision.
Inaccuracies in the vehicle localization, e.g. through a GPS position, would lead to inaccuracies in the correspondingly selected street map area.
In our experiments we employed the widely used Cityscapes~\cite{cordts2016cityscapes} dataset, for which we observed such inaccuracies.
Apart from this dataset, the currently most used semantic segmentation datasets in relevant publications do not contain all the required vehicle localization data in terms of latitude, longitude and heading.

We thus used the Cityscapes dataset and avoid GPS localization errors via an automatic correction algorithm
applied to the position, in order to demonstrate our approach.
Nevertheless, modern and future technologies like landmark detection and priors can provide a localization within a few centimeters~\cite{wilbers2019localization} and in real-world applications of our street map based validation approach, data with such precise localization information should be used.

\begin{figure}[t]
    \begin{subfigure}[c]{0.45\columnwidth}
    \centering
    \includegraphics[width=1.0\columnwidth]{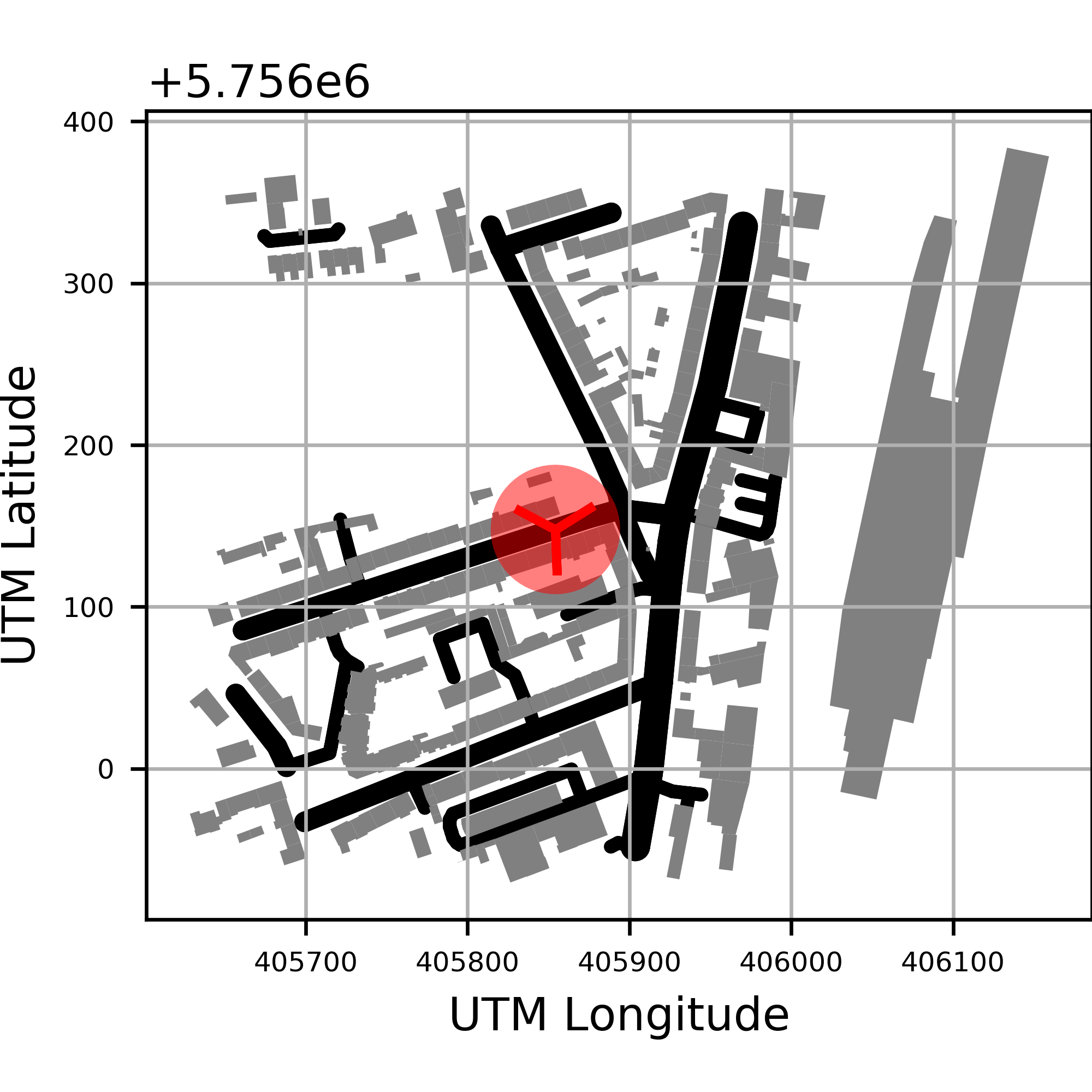}
    \subcaption{
    \textbf{Position range}
    }
    \label{fig:gpscorr_range}
    \end{subfigure}
    \begin{subfigure}[c]{0.55\columnwidth}
    \centering
    \includegraphics[width=1.0\columnwidth]{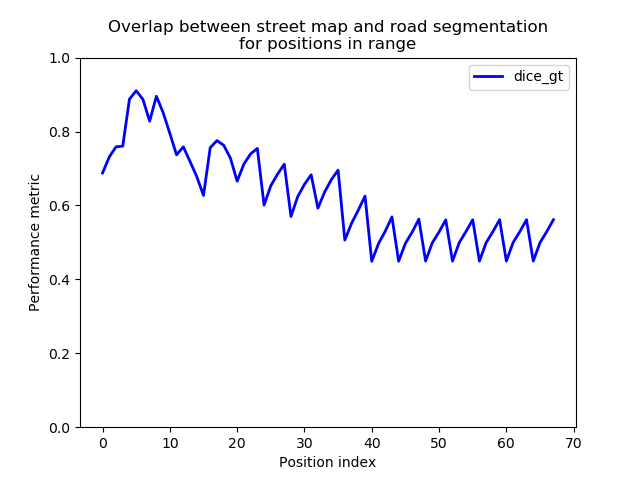}
    \subcaption{
    \textbf{Metrics for position range}
    }
    \label{fig:gpscorr_metrics}
    \end{subfigure}
    \begin{subfigure}[c]{1.0\columnwidth}
    \centering
    \includegraphics[width=1.0\columnwidth]{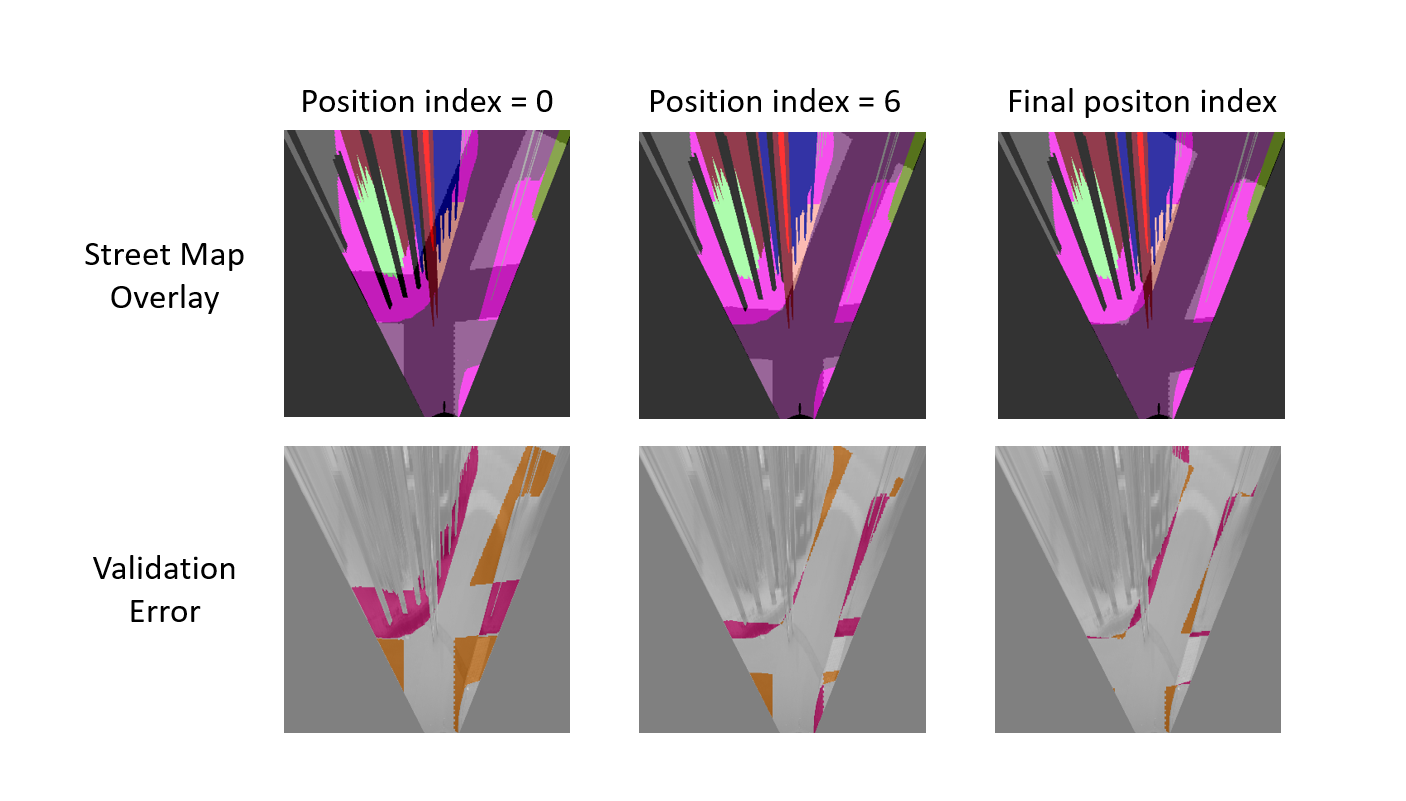}
    \subcaption{
    \textbf{Overlay and error for selected iterations}
    }
    \label{fig:gpscorr_ex}
    \end{subfigure}
\caption{
\textbf{Localization correction.}
This figure illustrates our algorithm for a localization correction based on ground-truth segmentation masks.
The algorithm takes a range around the original GPS position into account (a) and computes the overlap for potential positions that lie on roads within this range (b).
The saw tooth pattern in the metric diagram is the result from scanning the road not only along its length, but also along its width.
Figure (c) shows the overlap for specific iterations of positions and illustrates how the search algorithm finds the optimal position.
}
\label{fig:gpscorr}
\end{figure}

% --- correction/calibration according to G.T.
We developed an algorithm that corrects the GPS position based on an optimal fit with the ground truth segmentation. While in the predicted segmentation larger errors with respect to the street map can be expected (and it is our goal to identify them with our validation approach), the ground truth segmentation should have no or only small errors with respect to the street map. 
Using this assumption, the position can be calibrated according to the ground truth segmentation and can then be re-used for the street map based validation of the predicted segmentation.

% --- Algorithm
The goal of our algorithm is to identify the most accurate position within a GPS error range of a few metres for which the overlap between the street map and the ground truth segmentation is maximal.
Our algorithm is consists of three steps.
First, the position range according to the GPS error around the original position is determined. The road elements that lie in this range according to the street map are identified (see Figure~\ref{fig:gpscorr_range}).
The second step executes an optimization algorithm over a position search space: The identified road elements are rasterized into a grid of positions that lie along the length and the width of the road. The position headings are set to the angle of the corresponding road element. For each position in this search space, the dice coefficient, as defined in Section~\ref{sec:validationmetrics}, is computed (see Figure~\ref{fig:gpscorr_metrics}). The higher the dice coefficient, the better the fit of the ground truth segmentation to the street map and thus it is more probable that the position is the true position (see Figure~\ref{fig:gpscorr_ex}). The position with the maximum dice coefficient is saved as the new position.
The third step executes an additional optimization algorithm over a \textit{fine} position search space that covers the positions between the new position and the closest other positions from the previous search space. This step refines the found new position by applying a similar search routine as before, but now for the finer grid of positions.
\section{Experimental results}
\label{sec:experiments}

In this section we show comprehensive, statistical results from applying our approach to the Cityscapes segmentation dataset.
We present results on three aspects:
First, we apply our localization correction algorithm to the ground truth segmentations in order to find the most accurate GPS positions.
Using the corrected positions, we then apply our street map based validation approach on the predicted segmentations in order to identify potential prediction errors. 
Finally, we compare our approach to a validation of the predictions with ground truth data.

For our experiments, we use the Cityscapes train and validation subsets, for which ground truth segmentations are available and which comprise a total number of 3475 traffic scenes. We apply an additional data cleaning step and remove traffic scenes that contain segments with the label \textit{ground}, which describes areas that cars and pedestrians share equally. This label can not be assigned to either \textit{road} or no-\textit{road}, which is strictly relevant for our approach.

\subsection{Localization correction}

\begin{figure}[ht]
    \centering
    \includegraphics[width=1.0\columnwidth]{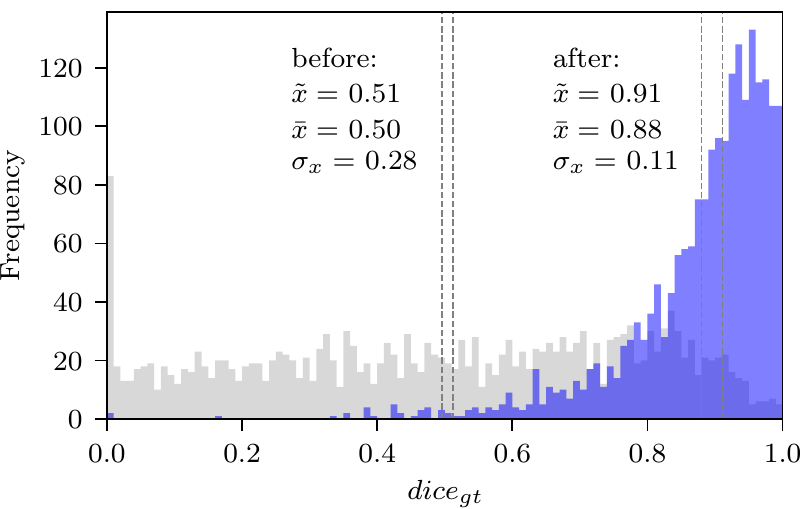}
\caption{
\textbf{Improvement by localization correction.}
Our algorithm optimizes the overlap between street map and ground truth segmentation, which is quantified by the $dice$ coefficient.
While it originally follows an equal distribution, which indicates a bad fit, after the correction it has a clear maximum at high values, which indicates a strongly improved and good fit.
}
\label{fig:gpscorr_dicehist}
\end{figure}

The localization correction significantly improves the accuracy of the GPS positions,
as shown in the distribution of the $dice$ coefficients before and after the correction in Figure~\ref{fig:gpscorr_dicehist}.
For the Cityscapes dataset we achieve an improvement from initially $dice = 0.50 \pm 0.28$, to $dice = 0.88 \pm 0.11$. The found GPS positions are saved for re-use in the validation of the predicted segmentations.

Although our algorithm significantly improves the fit between street map and ground truth segmentation, some segmentations cannot be sufficiently aligned.
A reason for this is the partially insufficient precision and information density of the used map itself.
Although open map providers like OpenStreetMap integrate content from various contributors, which can easily be updated, such sources often lack details. These could include, for example,
the precise geometric curvatures or the exact width of a road.
To alleviate such deficiencies, we apply a further data cleaning step and filter out the traffic scenes with a mediocre fit between ground truth segmentation and street map. We continue our analysis with the data subset above the median, i.e. for which $dice > 0.91$.

\subsection{Street map based segmentation validation}

\begin{figure}[t]
\centering
\includegraphics[width=\columnwidth]{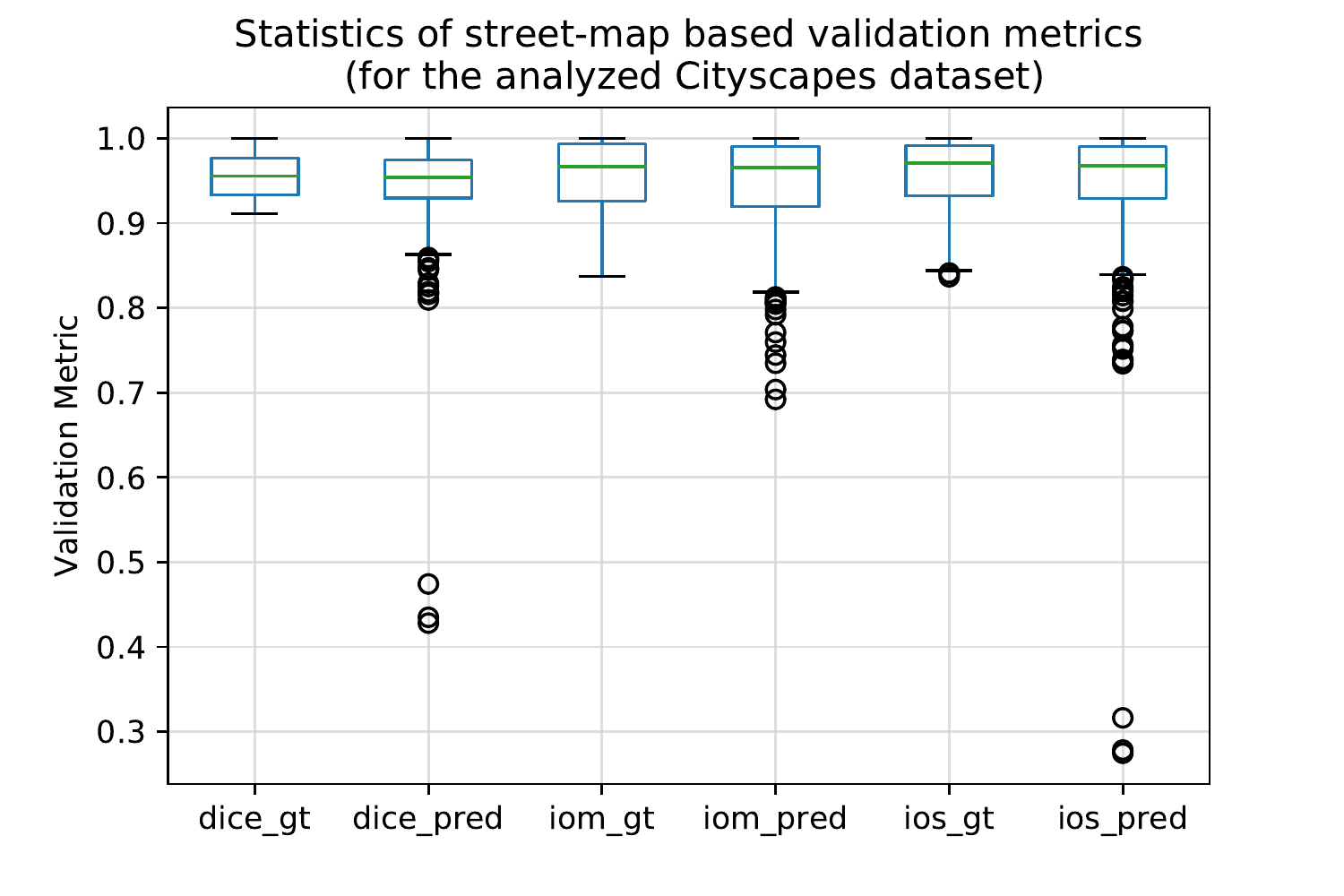}
\caption{
\textbf{Statistical summary of validation metrics.}
The box-whisker plots show the metrics $dice$, $IoM$, and $IoS$,
each for the ground truth and the predicted segmentation masks,
for the analyzed Cityscapes dataset.
The metrics for the predicted masks show clear outliers below the lower whiskers. These are indicators for validation errors, which we want to identify.
}
\label{fig:experiments_boxwhisker}
\end{figure}

We apply our street map based validation approach in order to identify those segmentations that contain potential prediction errors. Figure~\ref{fig:experiments_boxwhisker} shows the statistical summary of our validation metrics. All in all the metrics have high values of around 95\%, reflecting a general large overlap between the road segmentation and the street map. However, for the predicted segmentation there are some outliers that indicate validation errors. The existence of these outliers is expected and shows the functionality of our validation metrics.

\begin{figure}[t]
\centering
\includegraphics[width=\columnwidth]{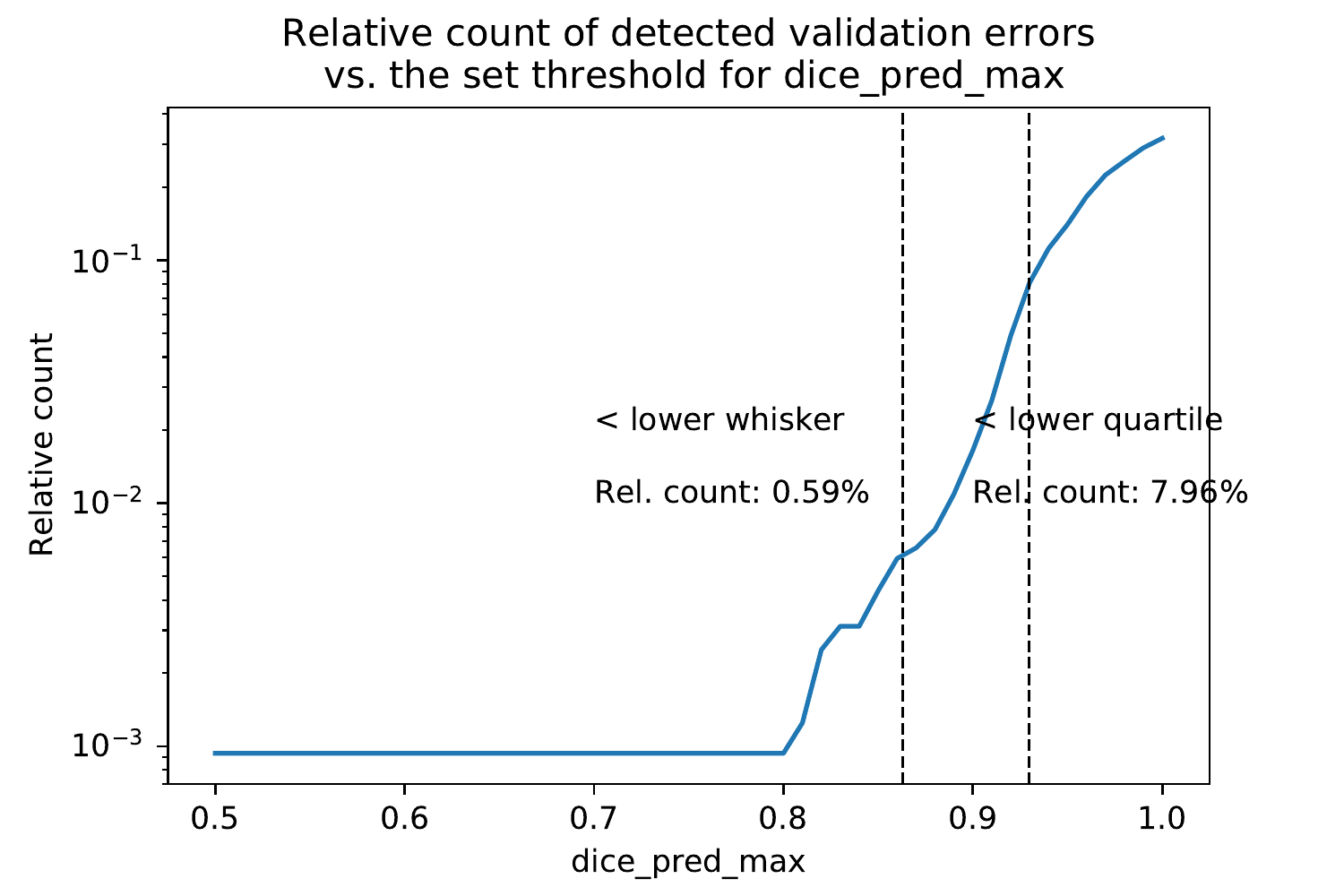}
\caption{
\textbf{Relative count of identified validation errors.}
%The quantity of identified validation errors depends on the set threshold for $dice_{\text{pred, max}}$.
The higher the threshold $dice_{pred, max}$, the larger the identification quantity.
Predicted segmentations that have a $dice$ coefficient below are counted as an validation error.
The highlighted values for the lower whisker and lower quartile refer to the thresholds deduced from the statistical summary in Figure~\ref{fig:experiments_boxwhisker}.
}
\label{fig:experiments_relcount}
\end{figure}

We identify the validation errors with the following procedure. First, we query all predicted segmentations that have a $dice$ coefficient below a specified threshold, which we call $dice_{\text{pred, max}}$. The set threshold defines the relative count of returned segmentations, as shown in Figure~\ref{fig:experiments_relcount}. For each returned instance, we further determine if it indicates a false positive or false negative error, depending on its $IoS$ and $IoM$ validation metrics.
Exemplary results for identified prediction errors are shown in the left columns of Figure~\ref{fig:experiments}.

\subsection{Comparison to ground-truth based validation}

In order to analyze the quality of the identified validation errors in the predicted segmentations, we compare them to the true prediction errors, which are the errors resulting from a validation using the ground truth segmentation.
We use two pixel-based measures to quantify the performance of our method. We compute 
the recall, which is the probability of detection, i.e., the probability that our method indicates a validation error, given a true error. Moreover, we compute the precision, which describes the probability of correctness, i.e., the probability that there is a true error, given that our method indicates a validation error.

\begin{figure*}[t]
    \begin{subfigure}[c]{1.0\textwidth}
    \centering
    \includegraphics[width=1.0\textwidth]{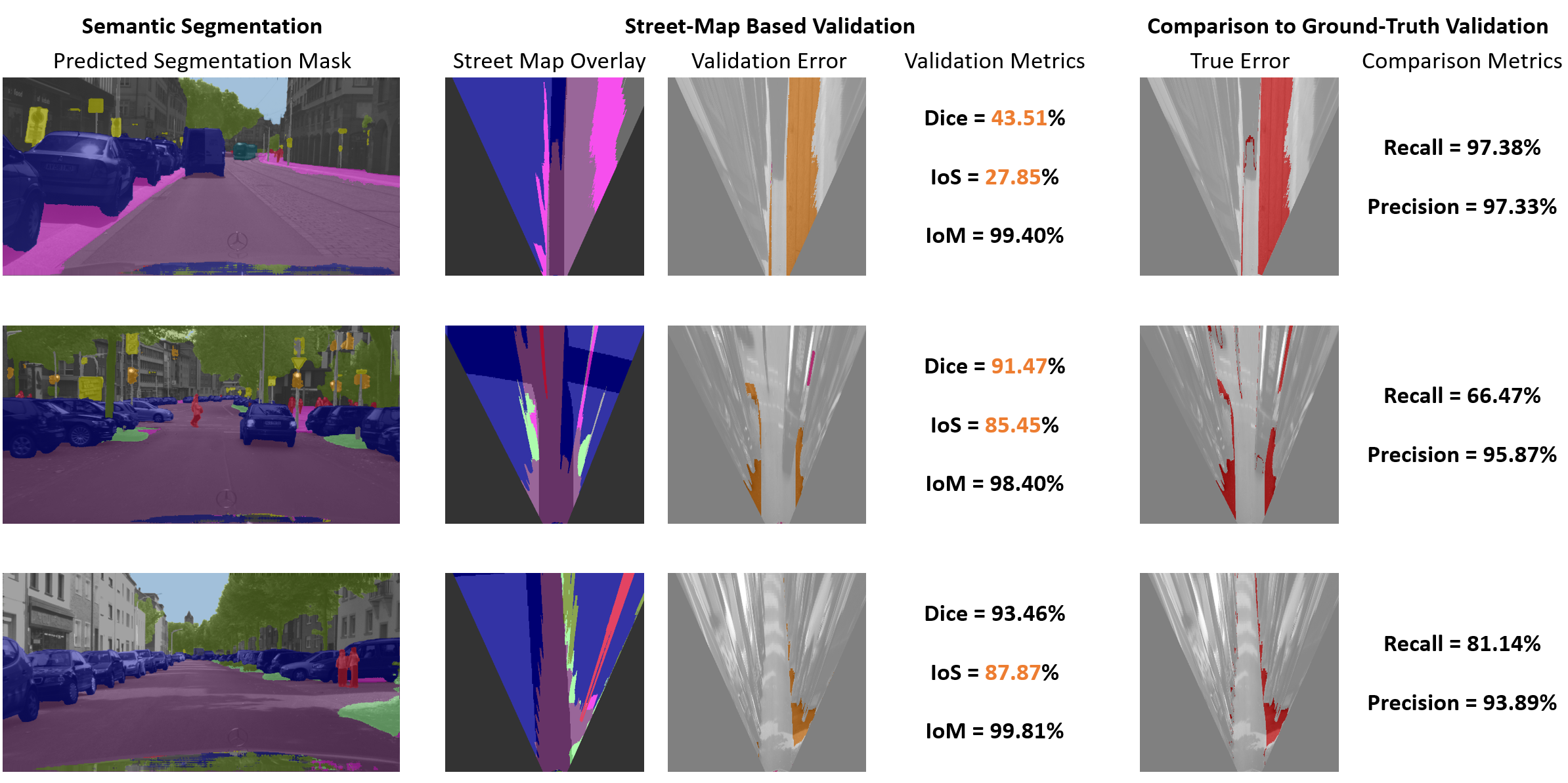}
    \subcaption{
    \textbf{Detection of false positives.}
    A low $IoS$ is an indicator for a false positive road segment in the predicted segmentation.
    }
    \label{fig:experiments_fp}
    \end{subfigure}
    \begin{subfigure}[c]{1.0\textwidth}
    \centering
    \includegraphics[width=1.0\textwidth]{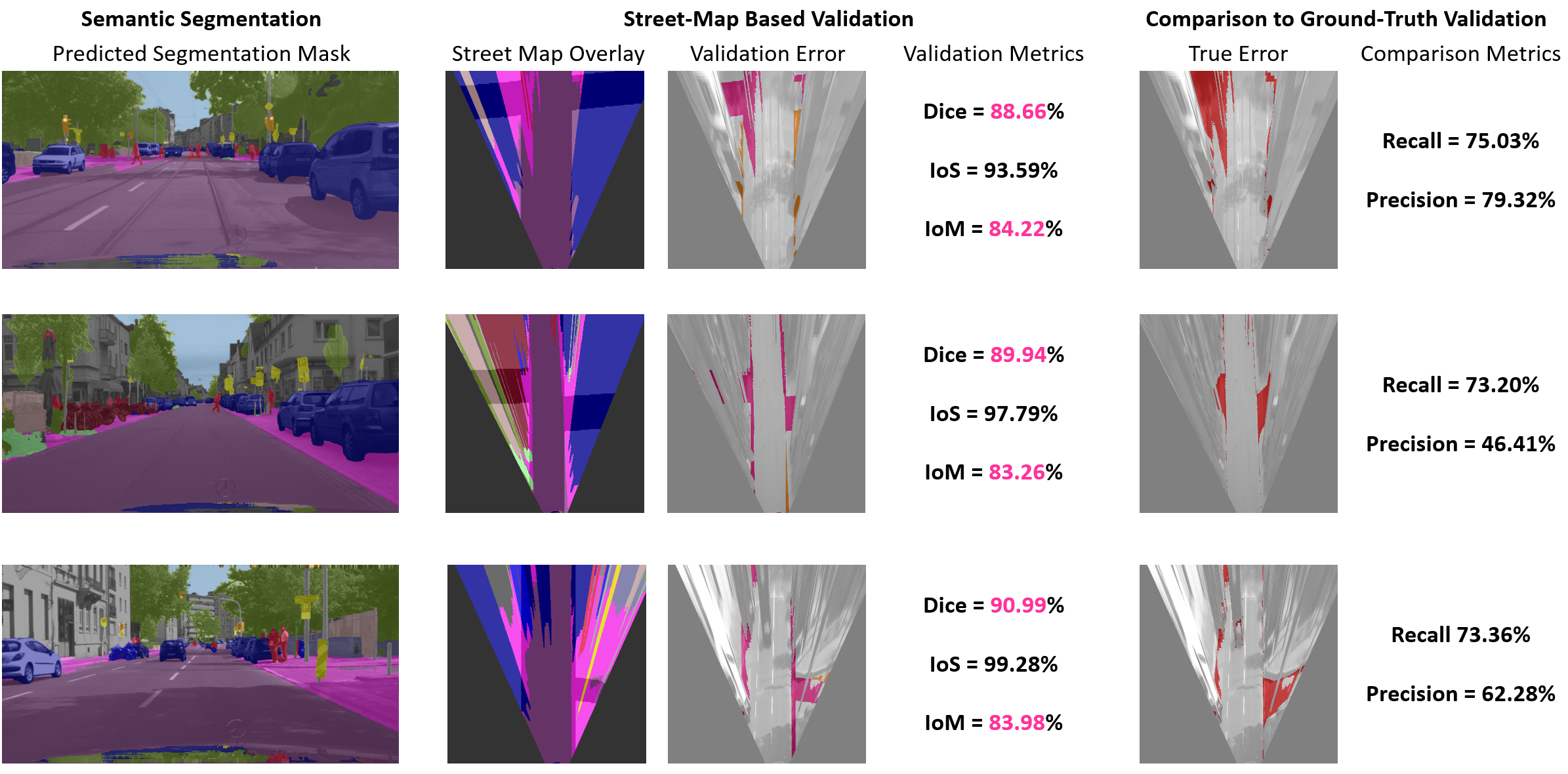}
    \subcaption{
    \textbf{Detection of false negatives.}
    A low $IoM$ is an indicator for a false negative road segment in the predicted segmentation.
    }
    \label{fig:experiments_fn}
    \end{subfigure}
\caption{
\textbf{Examples of experimental results for detected errors using our street map based validation.}
This figure shows how our street-map based validation approach can help to identify prediction errors. The rows show results from the validation error list, which is the automated output from our algorithm. The six rows here are a manual selection of that automated output.
Figure (a) shows exemplary results for detected false positives, and (b) for detected false negatives. Next to the actual error detection, we also show in the right columns how our method compares to a ground-truth based validation. It shows that the error regions are overlapping and the precision and recall give high values. This means that our validation method using street maps can identify similar error regions as a validation using ground truth data.
}
\label{fig:experiments}
\end{figure*}

Figure~\ref{fig:experiments} lists the recall and precision for exemplary results. It also visualizes the pixel regions from the identified validation error and the true error. As the regions mainly overlap, this shows that our method can uncover similar errors as a validation using ground truth.

\begin{figure}[t]
    \centering
    \includegraphics[width=\columnwidth]{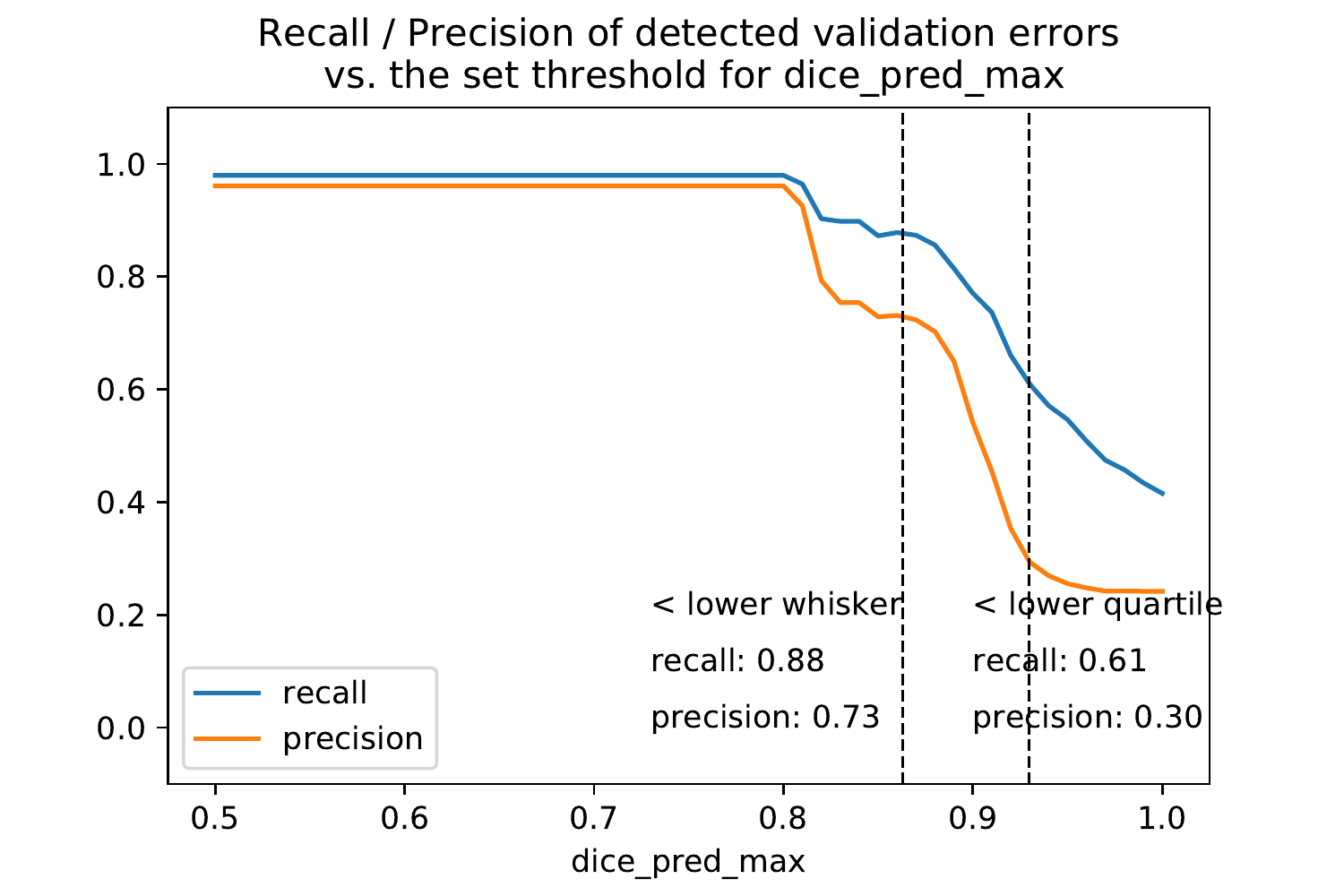}
    \caption{
    \textbf{Recall and precision of identified validation errors.}
    The lower the threshold $dice_{pred, max}$, the higher the identification quality in terms of recall and precision.
    The vertical lines highlight specific thresholds deduced from the statistical summary in Figure~\ref{fig:experiments_boxwhisker}.
    }
\label{fig:experiments_recprec}
\end{figure}

The recall and precision for the total analyzed dataset are shown in Figure~\ref{fig:experiments_recprec}. The smaller the selected threshold for filtering the outliers, the higher the precision and recall. However, a trade-off between quantity and quality of the found errors implies a medium threshold value, such as the lower quartile or whisker (based on the statistical summary in Figure~\ref{fig:experiments_boxwhisker}). Further analysis shows that precision and recall increase further with the size of the error region.

%\newpage
\section{Conclusion}

We proposed to validate machine learning models with a-priori domain knowledge and presented an approach that validates semantic segmentation masks with given street maps.

In particular, in our approach we combine the segmentation in bird's eye view with the field of view in the street map and use this overlay to compute validation errors.
We introduced two new validation metrics called Intersection over Segmentation ($IoS$) and Intersection over Map ($IoM$) that we used to identify segmentation masks with false positive and false negative road segments.
Furthermore, we developed an algorithm that can correct inaccurate vehicle positions by finding the best overlap with ground truth segmentation.
We performed an experimental study with the Cityscapes dataset and OpenStreetMap,
and showed that road segmentation errors can indeed be detected by our proposed validation procedure.

A general challenge is the precision of the street map itself. The validation procedure can only be as good as the information density in the map. Although our experiments with OpenStreetMap showed good results, even better results can achieved with high definition maps. For future work we therefore intend to perform the experiments with more precise and comprehensive maps, which would then also allow to investigate other classes than roads with our approach.

All in all, our proposed approach of a street map based validation of semantic segmentations offers a new and valuable way to support the goal of safe artificial intelligence in autonomous driving.

\section*{Acknowledgements}
Initial work has been funded by Volkswagen Group Automation (See Section~\ref{sec:approach} and~\cite{vonrueden2020towards}).
%Parts of this research have been
The remaining, extended elaboration of the work is 
funded by the Federal Ministry of Education and Research of Germany as part of the competence center for machine learning ML2R (01|S18038B).
%The authors gratefully acknowledges this support.

%\newpage
\bibliographystyle{IEEEtran}
\bibliography{references}

\begin{thebibliography}{10}
\providecommand{\url}[1]{#1}
\csname url@samestyle\endcsname
\providecommand{\newblock}{\relax}
\providecommand{\bibinfo}[2]{#2}
\providecommand{\BIBentrySTDinterwordspacing}{\spaceskip=0pt\relax}
\providecommand{\BIBentryALTinterwordstretchfactor}{4}
\providecommand{\BIBentryALTinterwordspacing}{\spaceskip=\fontdimen2\font plus
\BIBentryALTinterwordstretchfactor\fontdimen3\font minus
  \fontdimen4\font\relax}
\providecommand{\BIBforeignlanguage}[2]{{%
\expandafter\ifx\csname l@#1\endcsname\relax
\typeout{** WARNING: IEEEtran.bst: No hyphenation pattern has been}%
\typeout{** loaded for the language `#1'. Using the pattern for}%
\typeout{** the default language instead.}%
\else
\language=\csname l@#1\endcsname
\fi
#2}}
\providecommand{\BIBdecl}{\relax}
\BIBdecl

\bibitem{campbell2010autonomous}
M.~Campbell, M.~Egerstedt, J.~P. How, and R.~M. Murray, ``Autonomous driving in
  urban environments: approaches, lessons and challenges,'' \emph{Philosophical
  Transactions of the Royal Society A: Mathematical, Physical and Engineering
  Sciences}, vol. 368, no. 1928, 2010.

\bibitem{pendleton2017perception}
S.~D. Pendleton, H.~Andersen, X.~Du, X.~Shen, M.~Meghjani, Y.~H. Eng, D.~Rus,
  and M.~H. Ang, ``Perception, planning, control, and coordination for
  autonomous vehicles,'' \emph{Machines}, vol.~5, no.~1, 2017.

\bibitem{garcia2018survey}
A.~Garcia-Garcia, S.~Orts-Escolano, S.~Oprea, V.~Villena-Martinez,
  P.~Martinez-Gonzalez, and J.~Garcia-Rodriguez, ``A survey on deep learning
  techniques for image and video semantic segmentation,'' \emph{Applied Soft
  Computing}, vol.~70, 2018.

\bibitem{feng2019deep}
D.~Feng, C.~Haase-Schuetz, L.~Rosenbaum, H.~Hertlein, F.~Duffhauss, C.~Glaeser,
  W.~Wiesbeck, and K.~Dietmayer, ``Deep multi-modal object detection and
  semantic segmentation for autonomous driving: Datasets, methods, and
  challenges,'' \emph{arXiv:1902.07830}, 2019.

\bibitem{brundage2020toward}
M.~Brundage, S.~Avin, J.~Wang, H.~Belfield, G.~Krueger, G.~Hadfield, H.~Khlaaf,
  J.~Yang, H.~Toner, R.~Fong \emph{et~al.}, ``Toward trustworthy ai
  development: Mechanisms for supporting verifiable claims,''
  \emph{arXiv:2004.07213}, 2020.

\bibitem{burton2017making}
S.~Burton, L.~Gauerhof, and C.~Heinzemann, ``Making the case for safety of
  machine learning in highly automated driving,'' in \emph{International
  Conference on Computer Safety, Reliability, and Security}.\hskip 1em plus
  0.5em minus 0.4em\relax Springer, 2017.

\bibitem{kendall2017uncertainties}
A.~Kendall and Y.~Gal, ``What uncertainties do we need in {B}ayesian deep
  learning for computer vision?'' in \emph{Advances in Neural Information
  Processing Systems}, 2017.

\bibitem{mcallister2017concrete}
R.~McAllister, Y.~Gal, A.~Kendall, M.~Van Der~Wilk, A.~Shah, R.~Cipolla, and
  A.~V. Weller, ``Concrete problems for autonomous vehicle safety: Advantages
  of {B}ayesian deep learning.''\hskip 1em plus 0.5em minus 0.4em\relax
  International Joint Conferences on Artificial Intelligence, 2017.

\bibitem{cordts2016cityscapes}
M.~Cordts, M.~Omran, S.~Ramos, T.~Rehfeld, M.~Enzweiler, R.~Benenson,
  U.~Franke, S.~Roth, and B.~Schiele, ``The cityscapes dataset for semantic
  urban scene understanding,'' in \emph{Proceedings of the IEEE Conference on
  Computer Vision and Pattern Recognition}, 2016.

\bibitem{OpenStreetMap}
{OpenStreetMap}, \url{ https://www.openstreetmap.org }.

\bibitem{vonrueden2020informed}
L.~von Rueden, S.~Mayer, K.~Beckh, B.~Georgiev, S.~Giesselbach, R.~Heese,
  B.~Kirsch, J.~Pfrommer, A.~Pick, R.~Ramamurthy, M.~Walczak, J.~Garcke,
  C.~Bauckhage, and J.~Schuecker, ``Informed machine learning - a taxonomy and
  survey of integrating knowledge into learning systems,''
  \emph{arXiv:1903.12394v2}, 2020.

\bibitem{ardeshir2015geo}
S.~Ardeshir, K.~Malcolm Collins-Sibley, and M.~Shah, ``Geo-semantic
  segmentation,'' in \emph{Proceedings of the IEEE Conference on Computer
  Vision and Pattern Recognition}, 2015.

\bibitem{wang2015holistic}
S.~Wang, S.~Fidler, and R.~Urtasun, ``Holistic 3d scene understanding from a
  single geo-tagged image,'' in \emph{Proceedings of the IEEE Conference on
  Computer Vision and Pattern Recognition}, 2015.

\bibitem{diaz2016lifting}
R.~D{\'\i}az, M.~Lee, J.~Schubert, and C.~C. Fowlkes, ``Lifting gis maps into
  strong geometric context for scene understanding,'' in \emph{IEEE Winter
  Conference on Applications of Computer Vision}, 2016.

\bibitem{murthy2017shape}
J.~K. Murthy, S.~Sharma, and K.~M. Krishna, ``Shape priors for real-time
  monocular object localization in dynamic environments,'' in \emph{IEEE/RSJ
  International Conference on Intelligent Robots and Systems}, 2017.

\bibitem{schroeder2019using}
B.~Schroeder and A.~Alahi, ``Using a priori knowledge to improve scene
  understanding,'' in \emph{Proceedings of the IEEE Conference on Computer
  Vision and Pattern Recognition Workshops}, 2019.

\bibitem{xu2019spatial}
H.~Xu, C.~Jiang, X.~Liang, and Z.~Li, ``Spatial-aware graph relation network
  for large-scale object detection,'' in \emph{Proceedings of the IEEE
  Conference on Computer Vision and Pattern Recognition}, 2019.

\bibitem{tsai2018learning}
Y.-H. Tsai, W.-C. Hung, S.~Schulter, K.~Sohn, M.-H. Yang, and M.~Chandraker,
  ``Learning to adapt structured output space for semantic segmentation,'' in
  \emph{Proceedings of the IEEE Conference on Computer Vision and Pattern
  Recognition}, 2018.

\bibitem{wang2020train}
Y.~Wang, X.~Chen, Y.~You, L.~Erran, B.~Hariharan, M.~Campbell, K.~Q.
  Weinberger, and W.-L. Chao, ``Train in germany, test in the usa: Making 3d
  object detectors generalize,'' \emph{arXiv:2005.08139}, 2020.

\bibitem{romera2017erfnet}
E.~Romera, J.~M. Alvarez, L.~M. Bergasa, and R.~Arroyo, ``Erfnet: Efficient
  residual factorized convnet for real-time semantic segmentation,'' \emph{IEEE
  Transactions on Intelligent Transportation Systems}, vol.~19, no.~1, 2017.

\bibitem{wilbers2019localization}
D.~Wilbers, C.~Merfels, and C.~Stachniss, ``Localization with sliding window
  factor graphs on third-party maps for automated driving,'' in \emph{IEEE
  International Conference on Robotics and Automation}, 2019.

\bibitem{vonrueden2020towards}
L.~von Rueden, T.~Wirtz, F.~Hueger, J.~D. Schneider, and C.~Bauckhage,
  ``Towards map-based validation of semantic segmentation masks,''
  \emph{Workshop on AI for Autonomous Driving (AIAD) on the 37th International
  Conference on Machine Learning (ICML)}.

\end{thebibliography}

\end{document}